\documentclass[letterpaper, 10 pt, conference]{ieeeconf}  % Comment this line out if you need a4paper

\IEEEoverridecommandlockouts
\usepackage{graphics}
\usepackage{epsfig}
\usepackage{mathptmx}
\usepackage{times}
\usepackage{amsmath}
\usepackage{amssymb}
\usepackage{amsfonts}
\usepackage{array}
\usepackage{booktabs}
\usepackage{algorithm}
\usepackage{algpseudocode}
\usepackage{textcomp}
\usepackage{stfloats}
\usepackage{url}
\usepackage{graphicx}
\usepackage{balance}
\usepackage{cite}
\usepackage{multirow}
\usepackage{mathrsfs}
\usepackage[cal=cm]{mathalfa}

\title{\LARGE \bf
Geometric Reconstruction of Extrinsic Contact Trajectories \\using Tactile Sensing and Proprioception for Tool Manipulation
}

\author{Seojung Min$^{1}$, Yoonjin Kim$^{1}$, Jeong-Jung Kim$^{2}$ and Jung Kim$^{1}$% <-this % stops a space
\thanks{*This work was supported in part by the National Research Council of Science and Technology as part of the project titled “Development of core technologies for robot general purpose task artificial intelligence (RoGeTA) framework” under Grant NK261B.}% <-this % stops a space
\thanks{$^{1}$S. Min, Y. Kim and J. Kim are with Department of Mechanical Engineering, Korea Advanced Institute of Science and Technology, Daejeon 34141, Republic of Korea.
        {\tt\small seojung.min@kaist.ac.kr, yjk05@kaist.ac.kr, jungkim@kaist.ac.kr}}%
\thanks{$^{2}$J. J. Kim is with Department of AI Machinery, Korea Institute of Machinery and Materials, Daejeon 34141, Republic of Korea.
    {\tt\small rightcore@kimm.re.kr}}%
}

\begin{document}
\maketitle

\thispagestyle{empty}
\pagestyle{empty}

%%%%%%%%%%%%%%%%%%%%%%%%%%%%%%%%%%%%%%%%%%%%%%%%%%%%%%%%%%%%%%%%%%%%%%%%%%%%%%%%

\begin{abstract}
Tactile sensing enables robots to perceive rich contact information at the grasp, supporting tasks such as object recognition, in-hand pose estimation, and slip detection. 
However, in many tool-mediated manipulation tasks, the interaction that determines task success occurs at the tool tip, away from the tactile sensor, making direct sensing of tool-environment contact difficult, particularly when the contact moves during interaction.

In this work, we reconstruct the trajectory of extrinsic tool-tip contact using tactile sensing and robot proprioception. 
We formulate tool-tip trajectory reconstruction as a geometric inference problem under a single-point contact assumption. 
Our method first estimates the global tool-tip contact location from an anchoring segment designed to approximate fixed-point behavior, and then reconstructs the full trajectory by composing relative tool motion estimated from tactile marker observations under continuous contact.

Across $n=51$ trials with multiple trajectories, tools, wrist poses, and grasp configurations, the proposed pipeline achieves a trajectory RMSE of $8.59 \pm 2.41$\,mm in the world frame and a shape RMSE of $5.96 \pm 1.16$\,mm, while operating online at $14.00 \pm 4.11$\,Hz. 
Overall, the results show that extrinsic tool-tip trajectory geometry can be recovered consistently from grasp-level tactile sensing, with trajectory shape remaining stable across variations in tools, wrist poses, and grasp configurations.
\end{abstract}

%%%%%%%%%%%%%%%%%%%%%%%%%%%%%%%%%%%%%%%%%%%%%%%%%%%%%%%%%%%%%%%%%%%%%%%%%%%%%%%%
\begin{figure}[t]
    \centering
    \includegraphics[width=\columnwidth]{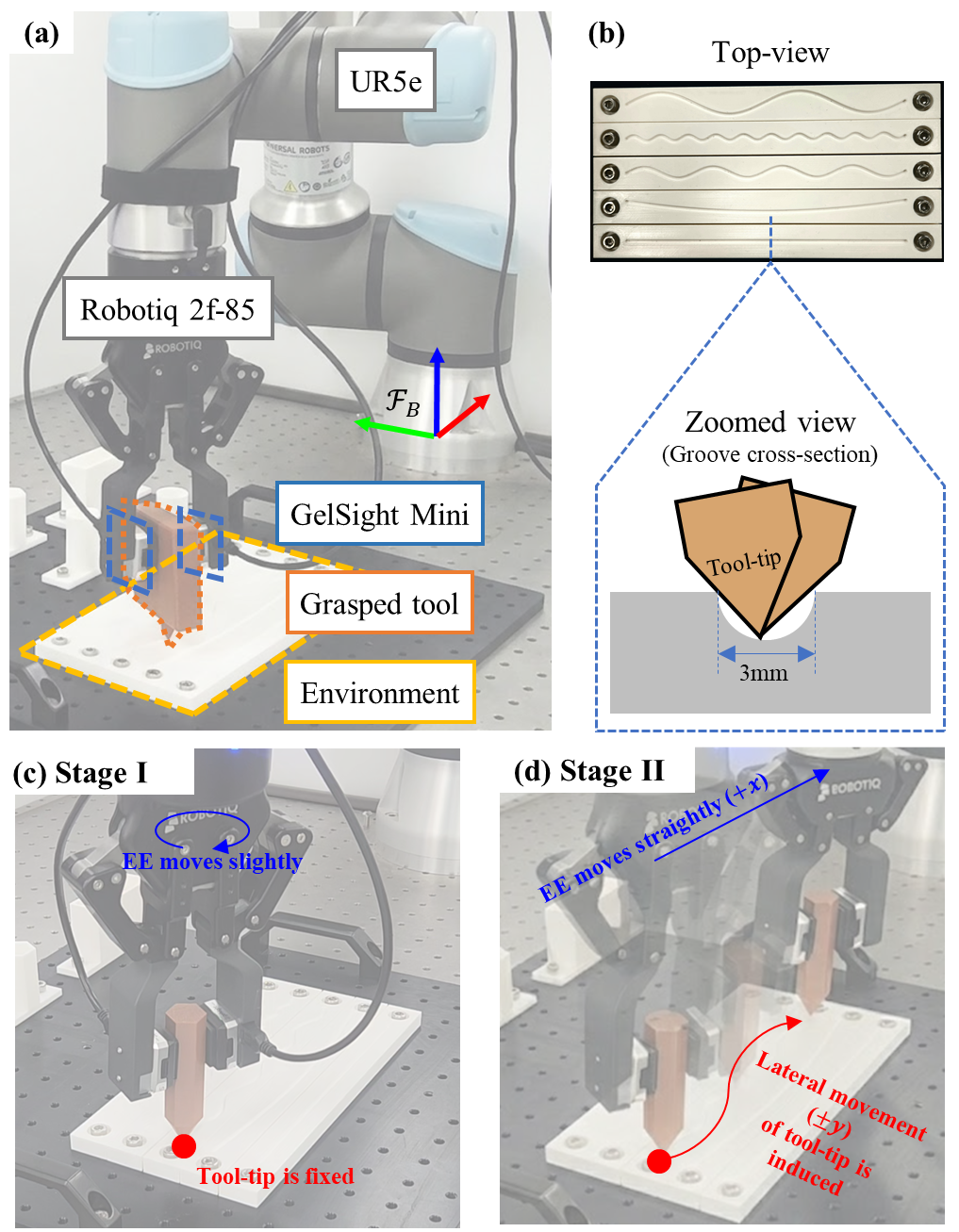}
    \caption{
    \textbf{Experimental setup and interaction scenario.}
    (a) Overall system setup,
    (b) Grooved structure defining the ground-truth trajectories (top view and cross-section),
    (c) Stage I: the end-effector moves slightly while the tool tip remains fixed, and
    (d) Stage II: the end-effector moves straight while the tool tip follows the groove.
    }
    \label{fig:system}
\end{figure}

%%%%%%%%%%%%%%%%%%%%%%%%%%%%%%%%%%%%%%%%%%%%%%%%%%%%%%%%%%%%%%%%%%%%%%%%%%%%%%%%

\begin{figure*}[t]
    \centering
    \includegraphics[width=\textwidth]{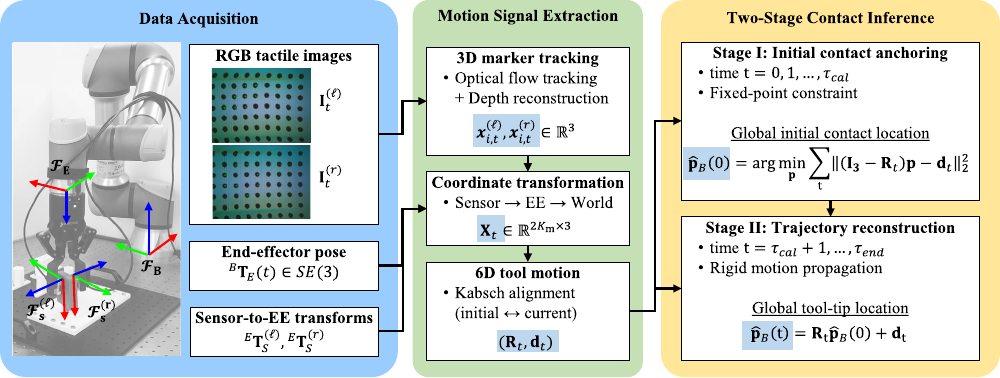}
    \caption{
    \textbf{Two-stage reconstruction pipeline.}
We estimate per-frame 6-DoF relative tool motion $(\mathbf R_t,\mathbf d_t)$ from dual tactile marker trajectories and anchor the initial tool-tip contact point $\hat{\mathbf p}_B(0)$ using a fixed-point anchoring segment (Stage I).
The full tool-tip trajectory $\hat{\mathbf p}_B(t)$ is then reconstructed by propagating the anchored point through the estimated motions (Stage II).
    }
    \label{fig:pipeline}
\end{figure*}

\section{Introduction}

Tactile sensing provides rich information about physical interaction and has become an important modality for robotic manipulation, particularly in situations where vision is unreliable due to occlusion or limited viewpoints. 
Recent vision-based tactile sensors such as GelSight~\cite{yuan2017gelsight}, DIGIT~\cite{lambeta2020digit}, TacTip~\cite{ward2018tactip}, GelSlim~\cite{taylor2022gelslim}, and 9DTact~\cite{lin20239dtact} enable high-resolution measurement of contact geometry, shear, and slip, supporting both learning-based and geometry-based manipulation approaches~\cite{luo2025tactile,li2025classification}.

Most prior work focuses on \emph{intrinsic tactile perception}, where contact occurs directly at the hand-object interface and the grasped object is the primary subject of interaction. 
Representative problems include object classification and material recognition from tactile cues\cite{spiers2016single,luo2016iterative,yuan2017shape,luo2018vitac}, in-hand pose estimation using tactile geometry and proprioception\cite{bimbo2016hand,villalonga2021tactile,bauza2023tac2pose,kelestemur2022tactile,yang2023hand}, and slip or shear detection for grasp stability and tactile servoing\cite{taylor2022gelslim,lloyd2024pose,ford2025shear,aquilina2024tactile}.

However, many real-world manipulation tasks are tool-mediated. 
In such tasks including writing, scraping, cutting, or insertion~\cite{shirai2023tactile,zhao2025tactile,higuera2023perceiving}, the grasp stabilizes a tool while task success depends on the interaction between the tool tip and the environment. 
This interaction constitutes an \emph{extrinsic contact} that occurs at a spatially separated location from the tactile sensor. 
As a result, the robot must infer the state of the tool-environment contact indirectly from deformation observed at the grasp.

A growing body of work has addressed extrinsic contact through state estimation and control. 
Ma et al.~\cite{ma2021extrinsic} introduced a geometric formulation for localizing extrinsic contact from tactile measurements under fixed-contact constraints, and subsequent work incorporated active control strategies for regulating contact configurations in manipulation tasks~\cite{kim2022active,kim2023simultaneous}. 
Extrinsic contact has also been used as feedback for manipulation skills such as pivoting, sliding, and non-prehensile manipulation~\cite{doshi2022manipulation,bronars2024texterity,oller2024tactile}. 
Learning-based approaches, such as Neural Contact Fields~\cite{higuera2022neural}, further estimate extrinsic contact distributions from tactile observations, while related studies consider joint contact-object estimation or extrinsic surface reconstruction~\cite{sipos2022simultaneous,van2026simultaneous,kim2025object}.

These approaches are complementary to our objective but solve a different problem. 
Fixed-contact formulations estimate contact from a motion segment in which the contact constraint remains unchanged; applying them independently along a moving trace would require repeated local anchoring rather than continuous trajectory recovery. 
Control-oriented methods regulate or exploit contact states for task execution, but the executed geometric trace of a remote tool tip is not their primary output.

In contrast to prior work that estimates extrinsic contact states or regulates contact for manipulation,
we reconstruct the \emph{continuous trajectory} of extrinsic tool-tip contact during tool-mediated interaction.
We formulate trajectory reconstruction as a geometric inference problem from tactile marker motion and robot proprioception under a single dominant point-contact assumption.
An initial anchoring segment estimates the global contact point by approximating fixed-point behavior in the world frame, and the full trajectory is then recovered by propagating this anchor through tactile-derived relative rigid motions.
The setup and pipeline are shown in Fig.~\ref{fig:system} and Fig.~\ref{fig:pipeline}.

We make the following contributions:
\begin{itemize}
\item We formulate extrinsic contact inference as \emph{trajectory reconstruction} from in-hand tactile sensing and proprioception.
\item We propose a two-stage geometric pipeline that uses fixed-point anchoring only for initialization and reconstructs the moving trace by propagating the anchored point with tactile-derived relative rigid motions.
\item We provide multi-condition experiments ($n=51$) and show that tactile-derived relative motion recovers trajectory shape consistently across the tested variations, while absolute localization is mainly limited by initial anchoring accuracy.
\end{itemize}

%%%%%%%%%%%%%%%%%%%%%%%%%%%%%%%%%%%%%%%%%%%%%%%%%%%%%%%%%%%%%%%%%%%%%%%%%%%%%%%%

\begin{figure*}[t]
    \centering
    \includegraphics[width=0.9\textwidth]{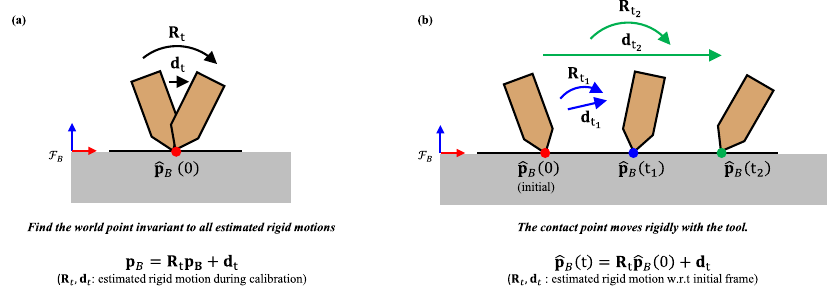}
    \caption{
    \textbf{Geometric principle of the two-stage reconstruction.}
        (a) \textbf{Stage I (anchoring):} during calibration, the tool tip is approximately stationary in the base/world frame, yielding fixed-point constraints that estimate the initial contact $\hat{\mathbf p}_B(0)$.
        (b) \textbf{Stage II (propagation):} the anchored point is propagated through the estimated relative rigid motions $(\mathbf R_t,\mathbf d_t)$ from tactile markers to reconstruct $\hat{\mathbf p}_B(t)$.
        }
    \label{fig:principle}
\end{figure*}

%%%%%%%%%%%%%%%%%%%%%%%%%%%%%%%%%%%%%%%%%%%%%%%%%%%%%%%%%%%%%%%%%%%%%%%%%%%%%%%%

\section{Methodology}
\label{sec:method}

Rather than estimating extrinsic contact as a single static state,
we reconstruct the temporal evolution of the tool-tip contact in the robot base frame.
The method has two stages.
\textbf{Stage I} anchors the initial tool-tip contact point using fixed-point constraints from an initial anchoring segment.
\textbf{Stage II} reconstructs the full trajectory by propagating this anchored point through tactile-derived relative rigid motions.
Algorithm~\ref{alg:pipeline} summarizes the pipeline.

\subsection{Problem Formulation}
\label{sec:problem_formulation}

\textbf{Frames:}
We use a left/right tactile image frame $\mathcal{F}_I^{(\ell)}$ and $\mathcal{F}_I^{(r)}$ (pixels),
corresponding tactile sensor frames $\mathcal{F}_S^{(\ell)}$ and $\mathcal{F}_S^{(r)}$,
a robot end-effector frame $\mathcal{F}_E$,
and the robot base frame $\mathcal{F}_B$ (world frame).

\textbf{Signals:}
At each time step $t$, we acquire dual tactile RGB images
$\mathbf{I}_t^{(\ell)}$ and $\mathbf{I}_t^{(r)}$
from the left/right sensors and the robot end-effector pose
${}^{B}\mathbf{T}_E(t)\in SE(3)$ from proprioception.

\textbf{Goal:}
Reconstruct the tool-tip trajectory $\hat{\mathbf p}_B(t)\in\mathbb{R}^3$ in the base frame
using only in-hand tactile sensing and robot proprioception.

\textbf{Assumptions:}
The tool is rigidly grasped throughout the interaction, with no slip between the tool and the elastomer pads of the tactile sensors.
The extrinsic interaction is dominated by a single effective point constraint at the tool tip.
We evaluate under passive contact constraints induced by a grooved environment, which maintains continuous extrinsic contact and restricts the tool-tip motion to a structured trajectory.
Each trial begins with an anchoring segment during which the tool tip is approximately stationary in the base frame (Sec.~\ref{sec:stage1}).

\textbf{Time indexing and trial structure:}
We consider a single interaction trial indexed by discrete frame index
$t \in \{0,1,\dots,\tau_{\mathrm{end}}\}$,
where $\tau_{\mathrm{end}}$ denotes the final frame index of the sequence.
Each trial consists of two consecutive segments:
an initial anchoring segment
$\mathcal A=\{0,1,\dots,\tau_{\mathrm{cal}}\}$,
during which the tool tip is approximately stationary in the base frame,
and a subsequent reconstruction segment
$\mathcal R=\{\tau_{\mathrm{cal}}+1,\dots,\tau_{\mathrm{end}}\}$,
during which the tool tip moves continuously under environment constraints.
Here, $\tau_{\mathrm{cal}}$ denotes the final frame index of the anchoring segment.

\subsection{Tactile Image Processing and Marker Tracking}
\label{sec:marker}

At each time step, we collect tactile RGB frames from dual tactile sensors (GelSight Mini) and track the motion of surface markers.
Assuming no relative slip between the tool and the elastomer pads,
the marker displacements correspond to a single rigid-body motion,
which we estimate as the tool motion (Sec.~\ref{sec:kabsch}).
Marker tracking and depth reconstruction are implemented based on the official GelSight Mini SDK,
with minor modifications and per-sensor calibration to ensure consistent metric scaling and depth normalization.

\textbf{Marker tracking (2D):}
For each sensor, marker centers are detected in the initial frame and tracked frame-to-frame
using pyramidal Lucas-Kanade (LK) optical flow implemented in OpenCV.
We use a window size of $15\times 15$, two pyramid levels,
and termination criteria of 10 iterations or $\epsilon=0.03$.
Tracking is initialized once at the beginning of each trial and maintained throughout the interaction.

\textbf{Depth reconstruction (3D lifting):}
For each tactile RGB frame, we reconstruct a dense depth map following the GelSight Mini SDK pipeline.
Specifically, (i) per-pixel surface normals are predicted using a lightweight neural network
from normalized RGB values and pixel coordinates,
and (ii) Poisson integration is applied to recover depth from the estimated gradients.
To account for sensor-specific offsets,
we subtract a pre-recorded baseline depth map for each sensor before converting depth values to metric units.
Marker pixels are excluded from normal estimation using an intensity-range mask.

\textbf{Depth-based marker selection:}
Not all tracked markers are equally informative due to partial contact and spatially non-uniform indentation.
At the beginning of each trial, we select $K_m=8$ informative markers
based on indentation depth after discarding the top-3 extreme-depth markers.
The deepest markers are excluded because they often lie in localized or saturated deformation regions that can violate the rigid-motion assumption.
The selected indices are fixed throughout the trial to maintain point correspondences.

\textbf{Output representation:}
For each sensor $s\in\{\ell,r\}$ and time step $t$,
the above procedure yields a fixed set of $K_m$ tracked markers with 3D coordinates in the sensor frame $\mathcal{F}_S^{(s)}$,

\begin{equation}
\mathbf x^{S^{(s)}}_{i,t} \in \mathbb{R}^3,\qquad i=1,\dots,K_m,
\end{equation}
which serves as the input to the base-frame transformation described in Sec.~\ref{sec:coord}.

%%%%%%%%%%%%%%%%%%%%%%%%%%%%%%%%%%%%%%%%%%%%%%%%%%%%%%%%%%%%%%%%%%%%%%%%%%%%%%%%
\begin{algorithm}[t]
\caption{Two-Stage Tool-Tip Trajectory Reconstruction}
\label{alg:pipeline}
\begin{algorithmic}[1]
\State \textbf{Input:} tactile images $\mathbf I_t^{(\ell)},\mathbf I_t^{(r)}$, end-effector pose ${}^{B}\mathbf T_E(t)$,
sensor-to-EE transforms ${}^{E}\mathbf T_S^{(\ell)},{}^{E}\mathbf T_S^{(r)}$
\State \textbf{Output:} reconstructed trajectory $\{\hat{\mathbf p}_B(t)\}_{t=0}^{\tau_{\mathrm{end}}}$

\State Detect markers and select $K_m$ markers per sensor at $t=0$

\For{$t = 0,1,\dots,\tau_{\mathrm{end}}$}
    \State Track markers and lift to 3D in sensor frames
    \State Transform to the base frame and stack $\mathbf X_t$
    \State Estimate relative motion $(\mathbf R_t,\mathbf d_t)$ from $\mathbf X_0$ to $\mathbf X_t$ 

    \If{$t \in \mathcal A$} \hfill (Stage I)
        \State Accumulate $(\mathbf I_3-\mathbf R_t)\mathbf p \approx \mathbf d_t$
    \ElsIf{$t = \min(\mathcal R)$} \hfill (Stage II)
        \State Solve $\hat{\mathbf p}_B(0)$ from accumulated constraints
        \State Propagate: $\hat{\mathbf p}_B(t)=\mathbf R_t\hat{\mathbf p}_B(0)+\mathbf d_t$
    \Else 
        \State Propagate: $\hat{\mathbf p}_B(t)=\mathbf R_t\hat{\mathbf p}_B(0)+\mathbf d_t$
    \EndIf
\EndFor
\end{algorithmic}
\end{algorithm}
%%%%%%%%%%%%%%%%%%%%%%%%%%%%%%%%%%%%%%%%%%%%%%%%%%%%%%%%%%%%%%%%%%%%%%%%%%%%%%%%

\subsection{Coordinate Transformation to the Base Frame}
\label{sec:coord}

Given the reconstructed 3D marker locations
$\mathbf x^{S^{(s)}}_{i,t}$ in each sensor frame,
we transform them to the robot base frame in two steps:
(i) sensor frame to end-effector frame,
and (ii) end-effector frame to base frame.

Each sensor is associated with a pre-calibrated rigid transform
to the robot end-effector frame,
denoted as ${}^{E}\mathbf T_{S}^{(\ell)}$ and ${}^{E}\mathbf T_{S}^{(r)}$.
These transforms are treated as constant throughout the experiments.

Using the robot proprioceptive pose ${}^{B}\mathbf T_E(t)$,
the base-frame marker coordinates are obtained as
\begin{equation}
\mathbf x^{B,(s)}_{i,t}
=
{}^{B}\mathbf T_E(t)
\, {}^{E}\mathbf T_{S}^{(s)}
\, \mathbf x^{S^{(s)}}_{i,t}.
\end{equation}

For each sensor, we stack the base-frame marker coordinates into
\begin{equation}
\mathbf X_t^{(s)}
=
\begin{bmatrix}
\mathbf x^{B,(s)}_{1,t} \\
\vdots \\
\mathbf x^{B,(s)}_{K_m,t}
\end{bmatrix}
\in \mathbb{R}^{K_m \times 3}.
\end{equation}

The full marker set used for rigid motion estimation is obtained
by concatenating the left and right sensor stacks:
\begin{equation}
\mathbf X_t
=
\begin{bmatrix}
\mathbf X_t^{(\ell)} \\
\mathbf X_t^{(r)}
\end{bmatrix}
\in \mathbb{R}^{N \times 3},
\quad N = 2K_m.
\end{equation}

where $N=2K_m$ is the fixed total number of selected markers from both sensors.

\subsection{6-DoF Relative Tool Motion Estimation via Kabsch}
\label{sec:kabsch}

Let $\mathbf X_0 \in \mathbb{R}^{N \times 3}$ denote the stacked
base-frame marker positions at the initial valid time step,
and $\mathbf X_t \in \mathbb{R}^{N \times 3}$ the marker positions
at time $t$, where $N = 2K_m$ is fixed across the sequence.

We estimate a relative rigid motion $(\mathbf R_t, \mathbf d_t)$
consisting of a rotation matrix $\mathbf R_t \in SO(3)$
and a translation vector $\mathbf d_t \in \mathbb{R}^3$
that maps the initial marker configuration to the current one by minimizing the pointwise alignment error.
\begin{equation}
(\mathbf R_t,\mathbf d_t)
=
\arg\min_{\mathbf R\in SO(3),\,\mathbf d\in\mathbb{R}^3}
\sum_{i=1}^{N}
\left\|
\mathbf R\,\mathbf x_{i,0}
+
\mathbf d
-
\mathbf x_{i,t}
\right\|_2^2 .
\end{equation}

This problem is solved using the Kabsch algorithm~\cite{kabsch1976solution},
which provides a closed-form SVD-based solution for optimal rigid alignment
when point correspondences are known.

The resulting $(\mathbf R_t,\mathbf d_t)$ represents the 6-DoF \emph{relative rigid motion} of the tool from the initial marker configuration to that at time $t$.
Note that $(\mathbf R_t,\mathbf d_t)$ is not an absolute pose in $\mathcal F_B$, but a motion estimate defined w.r.t.\ the initial valid frame.

\subsection{Stage I: Initial Contact Anchoring}
\label{sec:stage1}

Stage I estimates the initial tool-tip contact location
$\hat{\mathbf p}_B(0) \in \mathbb{R}^3$ in the robot base frame
using the anchoring segment $\mathcal A$ defined in Sec.~\ref{sec:problem_formulation}.
We adopt the fixed-point constraint formulation introduced by Ma et al.~\cite{ma2021extrinsic},
which models extrinsic contact under a stationary point assumption.
In this work, the formulation is used as an anchoring primitive
for subsequent trajectory reconstruction.

During an initial anchoring segment,
the tool tip is maintained approximately stationary
with respect to the base frame at an unknown location
$\mathbf p_B(0)$.
To induce observable tactile deformation while preserving this stationary contact,
we apply a small pre-planned motion that generates measurable marker displacement
without causing slip between the tool and the elastomer pads.

Given the estimated relative rigid motion
$(\mathbf R_t, \mathbf d_t)$ of the tool at time $t$
with respect to its initial configuration (Sec.~\ref{sec:kabsch}),
the fixed-point assumption implies
\begin{equation}
\mathbf p_B(0)
\approx
\mathbf R_t \mathbf p_B(0) + \mathbf d_t,
\qquad t \in \mathcal A.
\end{equation}

Rearranging yields
\begin{equation}
(\mathbf I_3 - \mathbf R_t)\mathbf p_B(0) \approx \mathbf d_t .
\end{equation}

Stacking these constraints over all anchoring frames
$t \in \mathcal A$
gives the linear least-squares problem
\begin{equation}
\hat{\mathbf p}_B(0)
=
\arg\min_{\mathbf p}
\left\|
\begin{bmatrix}
\mathbf I_3-\mathbf R_0\\
\vdots\\
\mathbf I_3-\mathbf R_{\tau_{\mathrm{cal}}}
\end{bmatrix}
\mathbf p
-
\begin{bmatrix}
\mathbf d_0\\
\vdots\\
\mathbf d_{{\tau_{\mathrm{cal}}}}
\end{bmatrix}
\right\|_2^2 .
\label{eq:fixedpoint_ls}
\end{equation}

\subsection{Stage II: Trajectory Reconstruction}
\label{sec:stage2}
Given the anchored initial contact location
$\hat{\mathbf p}_B(0)$ estimated in Stage I,
Stage II reconstructs the full tool-tip trajectory
over the reconstruction segment $t \in \mathcal R$
using the relative rigid motion estimates.

For each time step $t \in \mathcal R$,
the relative rigid motion $(\mathbf R_t, \mathbf d_t)$
estimated in Sec.~\ref{sec:kabsch}
represents the 6-DoF motion of the tool
with respect to its initial configuration.
Assuming rigid grasp without slip between the tool and the tactile pads,
the contact point evolves according to the same rigid motion.
Since $\mathbf x^{B}_{i,t}$ are expressed in the base frame, the relative motion $(\mathbf R_t,\mathbf d_t)$ maps base-frame coordinates of the initial configuration to base-frame coordinates at time $t$.

The reconstructed tool-tip trajectory is therefore given by
\begin{equation}
\hat{\mathbf p}_B(t)
=
\mathbf R_t \hat{\mathbf p}_B(0)
+
\mathbf d_t .
\label{eq:traj_recon}
\end{equation}

This formulation yields a trajectory-level estimate
of extrinsic contact motion,
rather than a per-frame contact state.
While Stage I provides a global anchoring of the contact point,
Stage II propagates this estimate temporally
using frame-wise rigid motion inferred from tactile marker deformation.
The resulting trajectory $\{\hat{\mathbf p}_B(t)\}_{t=0}^{\tau_\mathrm{end}}$
captures the continuous evolution of the tool-tip contact
in the base frame.

\begin{figure}[t]
    \centering
    \includegraphics[width=0.9\columnwidth]{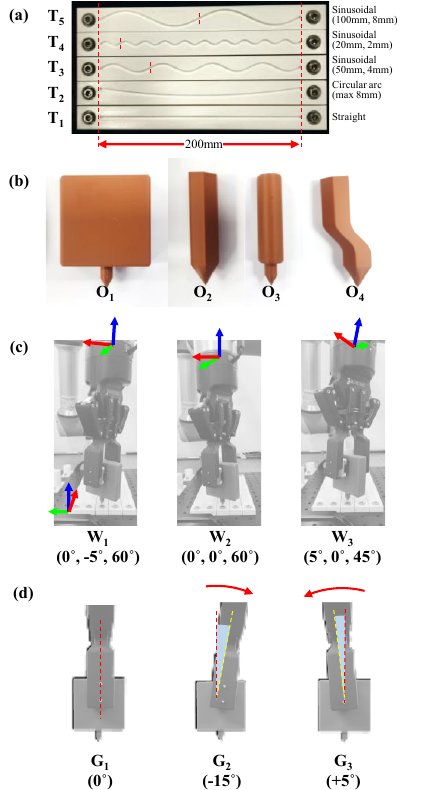}
    \caption{
    \textbf{Experimental conditions.}
    We evaluate consistency across variations in
    (a) trajectory (where two numbers of sinusoids denote period and amplitude, respectively),
    (b) tool geometry,
    (c) wrist pose (where the three angles stand for Euler angles in the robot base frame),
    and (d) grasp configuration.
    }
    \label{fig:conditions}
\end{figure}

%%%%%%%%%%%%%%%%%%%%%%%%%%%%%%%%%%%%%%%%%%%%%%%%%%%%%%%%%%%%%%%%%%%%%%%%%%%%%%%%
\begin{table}[t]
    \centering
    \caption{Summary of experimental condition sets.\\
    Each set varies one factor while keeping the others fixed.}
    \label{tab:experimental_conditions}
    \begin{tabular}{lcccc}
        \toprule
        \textbf{Set} & \textbf{Trajectory} & \textbf{Object} & \textbf{Wrist} & \textbf{Grasp} \\
        \midrule
        A & $T_1$--$T_5$ & $O_1$ & $W_1$ & $G_1$ \\
        B & $T_3$ & $O_1$--$O_4$ & $W_1$ & $G_1$ \\
        C & $T_3$ & $O_2$ & $W_1$--$W_3$ & $G_1$ \\
        D & $T_3$ & $O_1$ & $W_1$ & $G_1$--$G_3$ \\
        \bottomrule
    \end{tabular}
\end{table}

%%%%%%%%%%%%%%%%%%%%%%%%%%%%%%%%%%%%%%%%%%%%%%%%%%%%%%%%%%%%%%%%%%%%%%%%%%%%%%%%
\section{Experiments}
\label{sec:experiments}

We evaluate the proposed pipeline in a planar grooved environment that constrains the tool-tip motion.
The groove provides a structured and reproducible constraint for trajectory-level evaluation.

\begin{figure*}[t]
    \centering
    \includegraphics[width=0.95\textwidth]{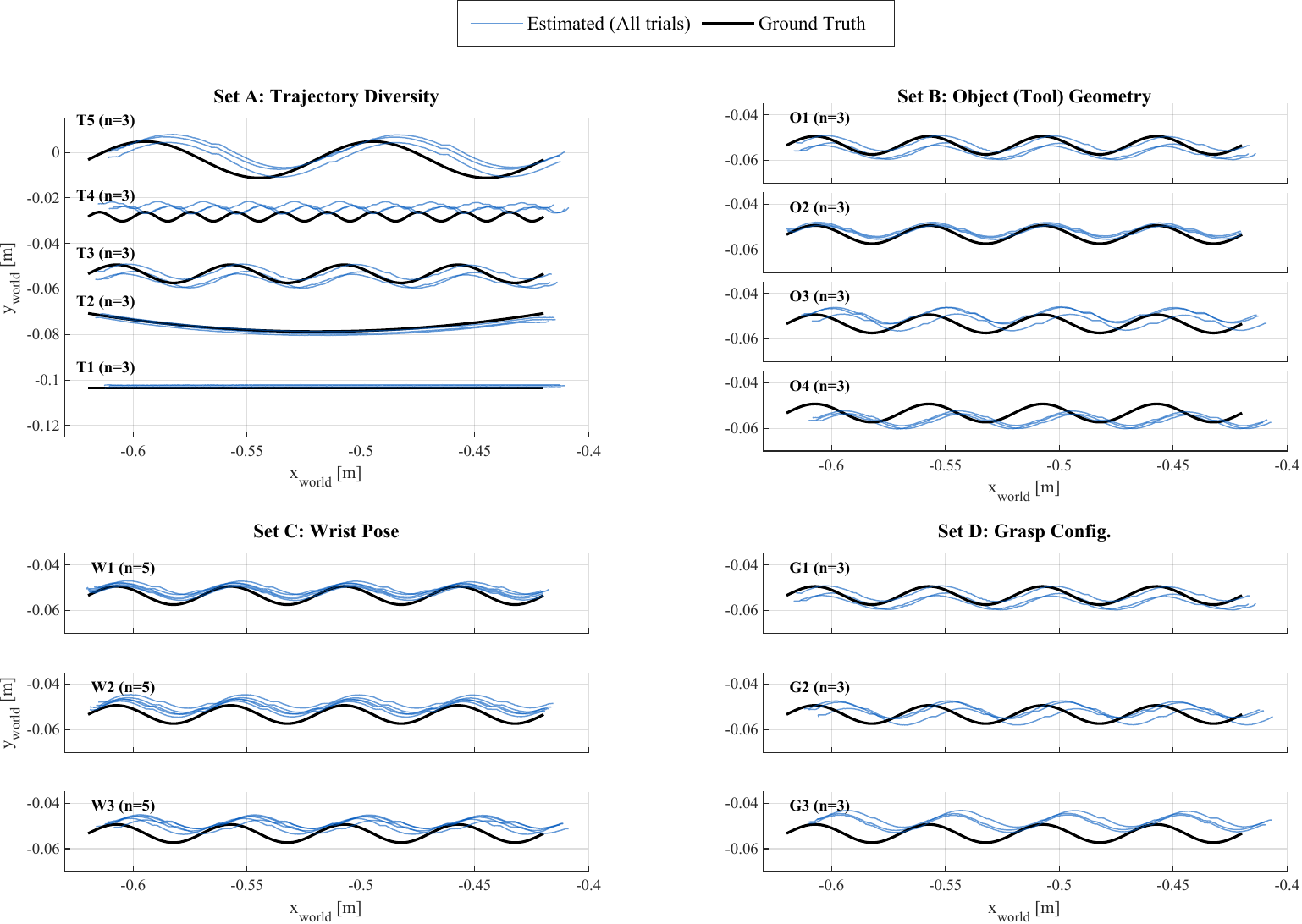}

    \caption{
        \textbf{Qualitative trajectory reconstruction.}
    Reconstructed trajectories (blue) are overlaid across trials and compared with ground-truth trajectories (black).
Each condition set varies one factor: trajectory shape (Set A), tool body geometry (Set B), wrist pose (Set C), or grasp configuration (Set D).
    }
    \label{fig:qualitative}
\end{figure*}
%%%%%%%%%%%%%%%%%%%%%%%%%%%%%%%%%%%%%%%%%%%%%%%%%%%%%%%%%%%%%%%%%%%%%%%%%%%%%%%%
\subsection{Experimental Setup}

\textbf{Robot and Control:}
A UR5e manipulator equipped with a Robotiq 2F-85 parallel gripper is used.
The robot operates under Cartesian position control using MoveIt-based planning.
Extrinsic contact is maintained purely by the groove constraint.

\textbf{Tactile Sensing:}
Two GelSight Mini sensors are mounted on the gripper fingers and used in all experiments.
RGB frames are resized to $320 \times 240$ for online processing.
We report the measured end-to-end loop rate and frames per trial rather than commanded speed.

\textbf{Tools:}
Four pencil-like tools ($O_1$--$O_4$) are evaluated, each 8.6\,cm in principal-axis length.
Cross-sections include rectangular, hexagonal, cylindrical, and a bent variant.
All tool tips are designed to fit a groove to promote approximate point-contact behavior.

\textbf{Groove and Trajectories:}
The grooved surface is fabricated via PLA 3D printing with a semicircular cross-section of 3\,mm diameter.
Five centerline trajectories are defined over a 200\,mm $x$-range: ($T_1$) straight line, ($T_2$) circular arc with maximum lateral deviation of 8\,mm, ($T_3$) sinusoid with period 50\,mm and amplitude 4\,mm, ($T_4$) sinusoid with period 20\,mm and amplitude 2\,mm, ($T_5$) sinusoid with period 100\,mm and amplitude 8\,mm.

\textbf{Interaction Protocol:}
For each trial, the robot grasps the tool with predefined wrist pose and grasp configuration.
The tool tip is aligned with the groove start point before execution.
During Stage I, a small pre-planned end-effector motion is applied while the groove constraint keeps the tool tip approximately fixed in the base frame.
During Stage II, the end-effector moves 20\,cm along the world-frame $+x$ direction under Cartesian control.
Although the commanded motion is purely along $x$, the groove geometry induces lateral tool-tip motion along $\pm y$ according to the trajectory shape.

\textbf{Condition Sets and Dataset Size:}

A total of $n=51$ trials are conducted across four condition sets (Table~\ref{tab:experimental_conditions}), varying trajectory, tool geometry, wrist pose, and grasp configuration.
The corresponding experimental variations are visually summarized in Fig.~\ref{fig:conditions}.

%%%%%%%%%%%%%%%%%%%%%%%%%%%%%%%%%%%%%%%%%%%%%%%%%%%%%%%%%%%%%%%%%%%%%%%%%%%%%%%%
\subsection{Ground Truth and Evaluation}
\label{sec:gt}

\textbf{Ground Truth:}
Ground truth is defined as the groove centerline trajectory in the robot base frame.
For each trajectory, the analytic curve is anchored at the measured starting point $\mathbf p^{gt}(0)$.
Since the tool tip conforms to the groove cross-section and promotes approximate point-contact behavior,
the centerline provides a physically meaningful approximation of the effective contact trajectory.

\textbf{Metrics:}
We report three complementary metrics (in millimeters):
(i) \textbf{trajectory RMSE (world frame)}
$\mathrm{RMSE}_{traj}=\mathrm{rms}(\|\hat{\mathbf p}_B(t)-\mathbf p^{gt}(t)\|_2)$,
(ii) \textbf{initial contact error}
$\|\hat{\mathbf p}_B(0)-\mathbf p^{gt}(0)\|$,
and (iii) \textbf{shape RMSE (initial-point aligned)},
where
$\hat{\mathbf p}_{shape}(t)=\hat{\mathbf p}_B(t)-\hat{\mathbf p}_B(0)+\mathbf p^{gt}(0)$
and
$\mathrm{RMSE}_{shape}=\mathrm{rms}(\|\hat{\mathbf p}_{shape}(t)-\mathbf p^{gt}(t)\|_2)$.

%%%%%%%%%%%%%%%%%%%%%%%%%%%%%%%%%%%%%%%%%%%%%%%%%%%%%%%%%%%%%%%%%%%%%%%%%%%%%%%%

\begin{table}[t]
\centering
\caption{Overall reconstruction performance ($n=51$).}
\label{tab:quantitative_summary}
\begin{tabular}{lc}
\toprule
\textbf{Metric} & \textbf{Mean $\pm$ Std} \\
\midrule
Trajectory RMSE (world frame) [mm] & $8.59 \pm 2.41$ \\
Shape RMSE (initial-point aligned) [mm] & $5.96 \pm 1.16$ \\
Initial contact error [mm] & $9.48 \pm 3.77$ \\
\midrule
Loop rate [Hz] & $14.00 \pm 4.11$ \\
\bottomrule
\end{tabular}
\end{table}

\section{Results}

\subsection{Reconstruction Accuracy and Error Structure}

Table~\ref{tab:quantitative_summary} summarizes reconstruction performance across all $n=51$ trials. 
The proposed method achieves a trajectory RMSE of $8.59 \pm 2.41$\,mm in the world frame.

Because this metric reflects both initialization error and trajectory propagation, we additionally evaluate trajectory shape accuracy after aligning the initial contact point. 
After this alignment, the shape RMSE decreases to $5.96 \pm 1.16$\,mm. 
This reduction indicates that a substantial portion of the world-frame trajectory error originates from bias in the initial anchoring stage rather than distortion of the reconstructed trajectory geometry.

Qualitative trajectory overlays shown in Fig.~\ref{fig:qualitative} illustrate this behavior.
Across trials, reconstructed trajectories follow the characteristic curvature patterns of the ground-truth groove trajectories.
For example, sinusoidal trajectories retain their periodic structure, and circular arc trajectories preserve their overall curvature.
In most cases, the main visible deviation between estimated and ground-truth trajectories appears as an approximately constant translational offset rather than progressive distortion along the trajectory.

To analyze the origin of this offset, we examine the relationship between anchoring accuracy and trajectory reconstruction error.
The initial contact error is measured as the Euclidean distance between the estimated initial contact point and the ground-truth trajectory origin.
Across all trials, the mean initial contact error is $9.48 \pm 3.77$\,mm.

\begin{figure}[t]
    \centering
    \includegraphics[width=0.95\columnwidth]{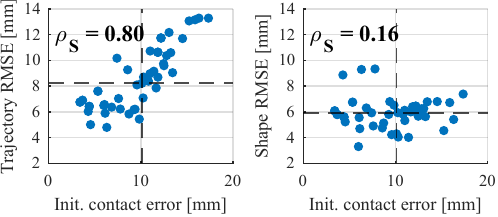}
    \caption{
\textbf{Error decomposition.}
Trajectory RMSE (world frame) correlates strongly with initial contact error, whereas shape RMSE shows weak correlation, indicating that global anchoring bias is a substantial contributor to absolute error.
    }
    \label{fig:error_analysis}
\end{figure}

Fig.~\ref{fig:error_analysis} shows the relationship between the initial contact error and trajectory reconstruction accuracy.
Trajectory RMSE exhibits a strong correlation with the initial contact error (Spearman's $\rho = 0.80$), whereas shape RMSE shows weak correlation with the same quantity (Spearman's $\rho = 0.16$).

This difference reveals a clear separation between two sources of reconstruction error.
Errors introduced during the anchoring stage primarily affect the absolute world-frame position of the reconstructed trajectory, while errors in relative motion estimation influence the trajectory geometry during propagation.

Together, these results indicate that while anchoring accuracy determines the absolute localization of the trajectory, the rigid-motion propagation stage recovers the relative geometry of the tool-tip motion with high consistency.

\subsection{Consistency Across Conditions}

\begin{figure}[t]
\centering
\includegraphics[width=\columnwidth]{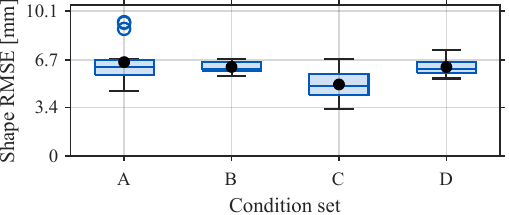}
\caption{
\textbf{Shape error across condition sets.}
Distribution of shape RMSE (initial-point aligned) for Sets A--D, corresponding to variations in trajectory shape, tool body geometry, wrist pose, and grasp configuration, respectively.
}
\label{fig:condition_shape}
\end{figure}

We analyze shape RMSE across the four condition sets in Table~\ref{tab:experimental_conditions}, which independently vary trajectory type, tool geometry, wrist pose, and grasp configuration.
As shown in Fig.~\ref{fig:condition_shape}, the mean shape error remains within $5.0$--$6.6$\,mm across all sets, indicating consistent shape reconstruction performance across the tested factors.
Both smooth and higher-frequency sinusoidal trajectories are reconstructed with comparable accuracy, and variations in tool geometry, wrist pose, and grasp configuration produce only minor differences.
Qualitative overlays in Fig.~\ref{fig:qualitative} further show no systematic failure across the tested condition sets.

\subsection{Runtime Performance}

The full reconstruction pipeline operates at $14.00 \pm 4.11$\,Hz (Table~\ref{tab:quantitative_summary}),
with observed loop rates ranging from $6.14$\,Hz to $23.68$\,Hz.

This runtime includes tactile image acquisition, marker tracking,
depth reconstruction, rigid motion estimation, and trajectory propagation
(Algorithm~\ref{alg:pipeline}). The majority of the computation time arises
from tactile image processing and depth reconstruction, while the geometric
estimation steps incur negligible overhead.

Despite these processing requirements, the loop rate remains sufficient for
continuous online trajectory reconstruction during tool-mediated interaction.
The pipeline processes tactile observations incrementally, allowing the
tool-tip trajectory to be updated as new measurements become available.
%%%%%%%%%%%%%%%%%%%%%%%%%%%%%%%%%%%%%%%%%%%%%%%%%%%%%%%%%%%%%%%%%%%%%%%%%%%%%%%%
%%%%%%%%%%%%%%%%%%%%%%%%%%%%%%%%%%%%%%%%%%%%%%%%%%%%%%%%%%%%%%%%%%%%%%%%%%%%%%%%
%%%%%%%%%%%%%%%%%%%%%%%%%%%%%%%%%%%%%%%%%%%%%%%%%%%%%%%%%%%%%%%%%%%%%%%%%%%%%%%%

\section{Discussion and Conclusion}

This work reframes extrinsic contact inference as a trajectory reconstruction problem and demonstrates that tool-tip trajectories can be recovered online from in-hand tactile sensing and robot proprioception. 
Across the tested trajectories, tools, wrist poses, and grasp configurations, the reconstructed paths preserve the characteristic geometry of the underlying grooves. 
The error analysis shows that world-frame trajectory error is strongly correlated with initial contact error, whereas initial-point-aligned shape error is weakly correlated with it. 
This suggests that a major error source is global anchoring offset, while tactile-derived relative motion remains informative for recovering trajectory geometry.

This observation suggests that reconstructed trajectory shape can be useful even when absolute anchoring is imperfect, particularly for tool-mediated interactions involving contour tracing or tactile exploration of surface features.

The present evaluation does not include a direct implementation baseline. 
A nominal proprioception-only baseline with a fixed tool transform would primarily follow the commanded straight end-effector motion and would not capture the lateral deviations induced by the groove. 
Therefore, the recovery of curved and sinusoidal trajectory components in our setup reflects information provided by tactile-derived relative motion. 
A calibrated tool-transform baseline could be useful for evaluating the initial anchoring stage in isolation, while active estimation-control frameworks provide a stronger comparison for closed-loop manipulation settings. 
We leave these comparisons to future work because they require either known tool-tip calibration or substantially different estimator-controller implementations.

A key limitation of the present study lies in the contact assumptions underlying the reconstruction model. The current formulation assumes a dominant single-point contact at the tool tip with continuous contact during interaction. To satisfy these assumptions and enable controlled evaluation, we use a grooved environment that passively constrains the tool tip and maintains stable contact throughout the motion. While this setup provides a clean trajectory-level ground truth and isolates the estimation problem, it does not represent more general manipulation scenarios where contact conditions may vary, slip may occur, or contact may be intermittent.

Future work will therefore focus on extending the framework beyond passive constraints by incorporating contact-maintenance strategies such as tactile servoing or force-based control. Rather than removing the underlying contact assumptions, such control strategies could actively regulate the interaction to maintain the conditions required by the reconstruction algorithm, enabling the same estimation framework to operate in less structured environments. Another promising direction is to use reconstructed tool-tip traces to infer local environment geometry during interaction, supporting tactile exploration and geometry estimation in tool-mediated manipulation.

\section*{ACKNOWLEDGMENT}

Portions of the manuscript were edited with the assistance of ChatGPT (OpenAI). All technical content, experiments, and conclusions were developed and verified by the authors.
\bibliographystyle{IEEEtran}
\bibliography{IEEEabrv, ref}

\end{document}